\documentclass{CVIS}
\usepackage{amsmath,bm}
\usepackage{amsfonts}
\usepackage[compact]{titlesec}  
\usepackage{graphicx}% Include figure files
\usepackage{dcolumn}% Align table columns on decimal point
\usepackage{subcaption,soul}
\usepackage{caption}
\usepackage[numbers,sort&compress,sectionbib]{natbib}
\usepackage[pagebackref=flase,breaklinks=true,letterpaper=true,colorlinks=true,linkcolor = blue, urlcolor = black,citecolor=black,bookmarks=true]{hyperref}
\usepackage{url}
\usepackage{mleftright}
\usepackage{multirow}

\usepackage{titlesec}
\usepackage[redeflists]{IEEEtrantools}
\titlespacing{\section}{0pt}{0pt}{0pt}

\title{NMR: Neural Manifold Representation for Autonomous Driving.}
\begin{document}
\setlength{\abovedisplayskip}{5pt}
\setlength{\belowdisplayskip}{5pt}
\setlength{\belowcaptionskip}{-10pt}
\bstctlcite{IEEEexample:BSTcontrol}
\sloppy
\author{
\begin{tabularx}{\textwidth}{X X}
Unnikrishnan R. Nair$^*$ & Ola Electric\\
Sarthak Sharma$^*$  & Ola Electric\\
Midhun S. Menon$^*$ & Ola Electric\\
Srikanth Vidapanakal & Ola Electric\\
\multicolumn{2}{l}{Email: \{unnikrishnan.r, sarthak.sharma1, midhun.s, srikanth.vidapanakal\}@olaelectric.com}
\end{tabularx}
\thanks{*Equal contribution.}
}

\maketitle
\begin{abstract}
Autonomous driving requires efficient reasoning about the Spatio-temporal nature of the semantics of the scene. Recent approaches have successfully amalgamated the traditional modular architecture of an autonomous driving stack comprising perception, prediction, and planning in an end-to-end trainable system. Such a system calls for a shared latent space embedding with interpretable intermediate trainable projected representation. One such successfully deployed representation is the Bird's-Eye View(BEV) representation of the scene in ego-frame. However, a fundamental assumption for an undistorted BEV is the local coplanarity of the world around the ego-vehicle. This assumption is highly restrictive, as roads, in general, do have gradients. The resulting distortions make path planning inefficient and incorrect. To overcome this limitation, we propose Neural Manifold Representation (NMR), a representation for the task of autonomous driving that learns to infer semantics and predict way-points on a manifold over a finite horizon, centered on the ego-vehicle. We do this using an iterative attention mechanism applied on a latent high dimensional embedding of surround monocular images and partial ego-vehicle state. This representation helps generate motion and behavior plans consistent with and cognizant of the surface geometry. We propose a sampling algorithm based on edge-adaptive coverage loss of BEV occupancy grid and associated guidance flow field to generate the surface manifold while incurring minimal computational overhead. We aim to test the efficacy of our approach on CARLA and SYNTHIA-SF.  
\end{abstract}
\section{Introduction}
Autonomous driving is one of the most active research areas of this decade.
% Thanks to huge market potential, cheap compute availability, cloud and GPU computing, and rapid advances in machine learning, this field shows promising results, garnering heavy funding and attracting more researchers. 
The operating environment for a self-driving system is highly complex, diverse, and dynamic. Navigating such scenes warrants reasoning and behavior planning that jointly reasons in the spatio-temporal domain of the scene. In this context, end-to-end systems with interpretable and trainable projections of shared latent representations have shown promising results. One exciting representation is the Bird's-eye View (BEV). BEV is a top-down view of the space around the ego-vehicle in an egocentric frame of reference. It is also the native space for path/behavior planning. Different groups have proposed multiple architectures \cite{mp3,p3,lss,fiery,nmp} in the recent past to derive the BEV of the scene from raw sensor inputs. However, the BEV assumes the coplanarity of the ego-vehicle and other agents in the scene. This assumption is highly restrictive, which, when relaxed, distorts the generated representation and makes it unintuitive for planning. We propose NMR, a surface representation amenable to the task of end-to-end autonomous driving on a non-planar road. We test it on a network that does waypoint prediction using a learnt Guidance Offset Field (GOF) and dense Semantic Occupancy Grid (SOG) prediction , which aims to predict semantic labels at any spatio-temporal query location $\displaystyle{(x, y, z, t)}$ on the manifold, bounded by some spatial range and the time interval. The sparse task of waypoint prediction is further aided by a dense prediction task of future semantic occupancy. We further improve the performance of the proposed approach by incorporating an attention based feature thresholding in the network. Finally, we improve the architecture's scalability by incorporating adaptive sampling based on edge distance transform and coverage loss, generating well-resolved segmentation maps without incurring a high computation cost.
\section{Problem Statement}
An ego-vehicle has to travel from point 1 to point 2 in a generic driving environment. The route for this is given in terms of sparse level target points. The driving environment includes jaywalkers, pedestrians, traffic lights and other vehicles. The driving surface can be a planar/ non-planar manifold. 

% \begin{figure}[!ht]
% \centering
%   \includegraphics[width=1.0\linewidth]{teaser_cropped_mod.jpeg}
%   \caption{\textbf{Visualization of a manifold representation for non-planar roads.} The grey bordered image shows a scene from Carla \cite{carla} with a steeply graded road. Most of the existing pipelines \cite{lss,fiery,neat,nmp} utilise a BEV representation as shown in the black bordered image for scene understanding and planning. However as we can see from images bordered with red, green and blue such a representation is not adequate for path planning. No Line-of-Sight(LoS) considerations have been incorporated. This can cause issues at inference due to information mismatch and BEV distortion. Also, gradient information of drivable surface is beneficial for planning. The NMR architecture infers the scene semantics and waypoints which are consistent with the surface as shown in the bottom most image.}
% \label{fig:teaser}
% \end{figure}
\begin{figure}[!ht]
\centering
  \includegraphics[width=1.0\linewidth]{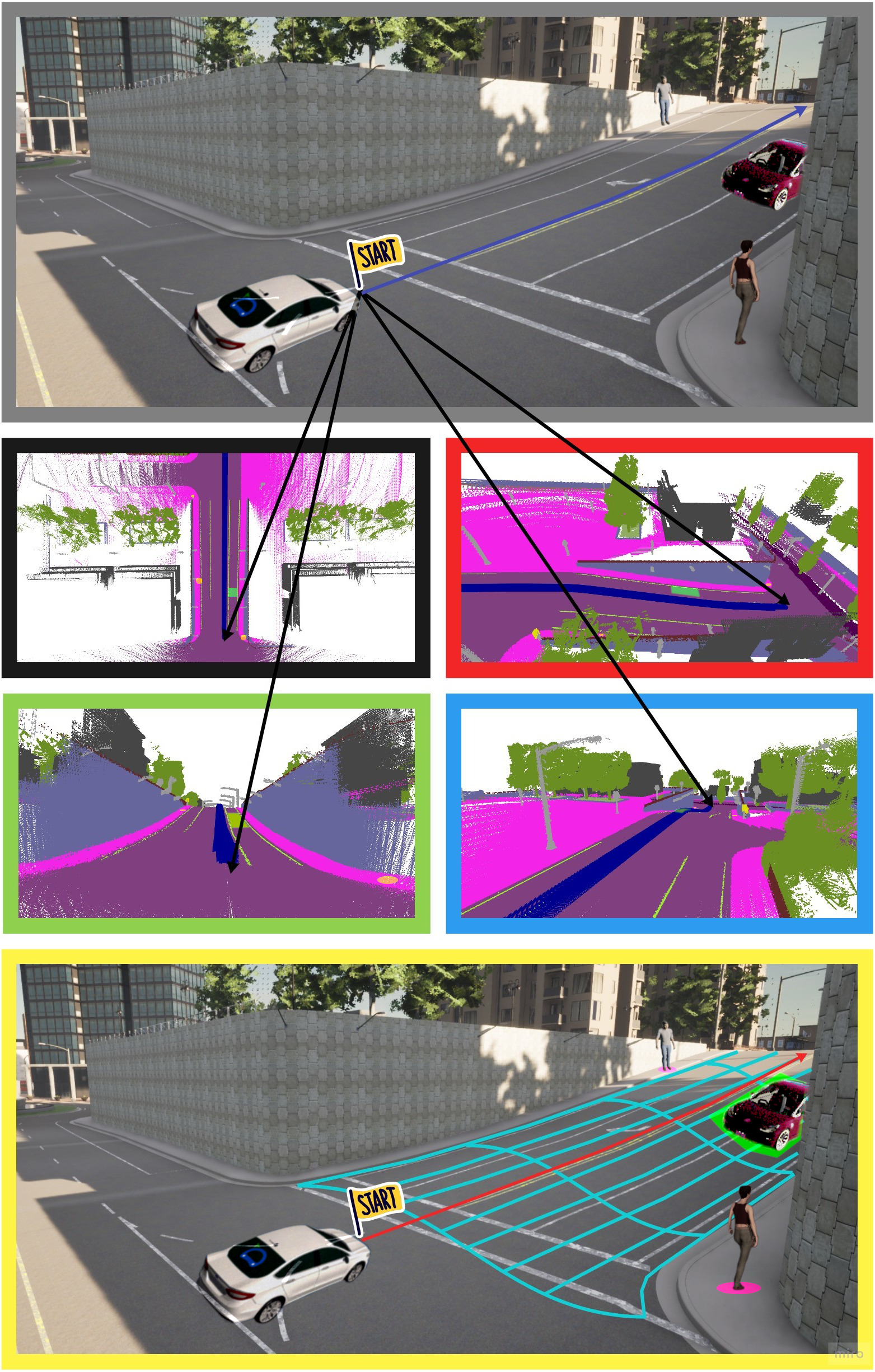}
  \caption{The image with grey borders shows a scene from Carla \cite{carla} with a steeply graded road. Notice the loss on information (Line-of-Sight, gradient information of surface) in BEV representation (black border) - as compared to the actual surface topology (red/blue/green border). The NMR architecture infers the scene semantics and waypoints which are consistent with the surface as shown in the bottom most image.}
\label{fig:teaser}
\end{figure}

\section{Methodology}

% \begin{figure*}[!ht]
%   \includegraphics[width=0.98\textwidth]{arch.png}
%   \caption{\textbf{Architecture}: The encoder takes in images from a surround monocular setup, and extract features via Res-Net \cite{resnet}. These image features along with lifted velocity features and learned positional embeddings are summed and fed into a transformer. An iterative attention field updates an attention map of the obtained latent embedding $\textbf{Z}$. In each iteration, the decoder predicts the control points $\{\textbf{P}_c^{ef}\}_{e=1\dots E,f=1\dots F}$ that define the B{\'e}zier representation of the road surface, along with semantics and offsets in the $(s_u,s_v)$ space given a query point $\textbf{q}$ on the manifold. At test time, we sample points from surface to obtain predictions for offsets and semantics}
% \label{fig:arch}
% \end{figure*}

% We attack this problem at two levels. Firstly, we propose NMR, a low dimensional representation for smooth surfaces. Secondly, we propose a network architecture motivated by NEAT\cite{neat}, but with key improvements in sampling methodology to improve the SOG resolution and quality without incurring much additional computational overheads.
% The Sec.\ref{sec:surf-rep} explains the surface representation, and Sec.\ref{sec:arch} discusses the network architecture.
\subsection{Surface Representation} \label{sec:surf-rep}
To represent surfaces smoothly, we propose NMR. NMR is a two-staged parametrization of the surface. In the first stage, the Cartesian points on the surface $(x,y,z)\in\mathbb{R}^{3}$ is mapped to parameters $(u, v)$ $\in \mathbb{R}^{2},~0\leq u,v \leq 1$. In the next stage, $(u, v)$ parametric space is transformed to the surface isometric arc length mapping $(s_u, s_v)$, giving intuitive two-dimensional mapping on the surface as it is topologically invariant. The two mappings when combined with any set of smooth basis functions provide an intuitive two-dimensional representation for a surface. In this work, without loss of generality, we choose the B{\'e}zier surface\cite{bezier1977essai} which uses Bernstein polynomial basis, denoted as $B(u)$. Hence any point on the surface $\textbf{P}\in\mathbb{R}^3$ is represented as linear combination of Bernstein polynomial product basis functions $B_{e}^E(u)\cdot B_{f}^F(v)$ and an $E \ast F$ net of control points $ \{P_c^{ef}\}_{e \in {1\dots E}, f \in {1 \dots F}}$, mapping from $0\leq u,v \leq 1\longmapsto (x,y,z)$. 
\begin{align}
\label{eqn:bspline}
\begin{split}
 \textbf{P} &= \sum_{e=1}^{E}\sum_{f=1}^{F} \textbf{P}_{c}^{ef} \cdot B_{e}^E(u)\cdot B_{f}^F(v) ,
\end{split}
\end{align}
\subsection{Network architecture} \label{sec:arch}
\begin{figure*}[!ht]
  \includegraphics[width=\textwidth]{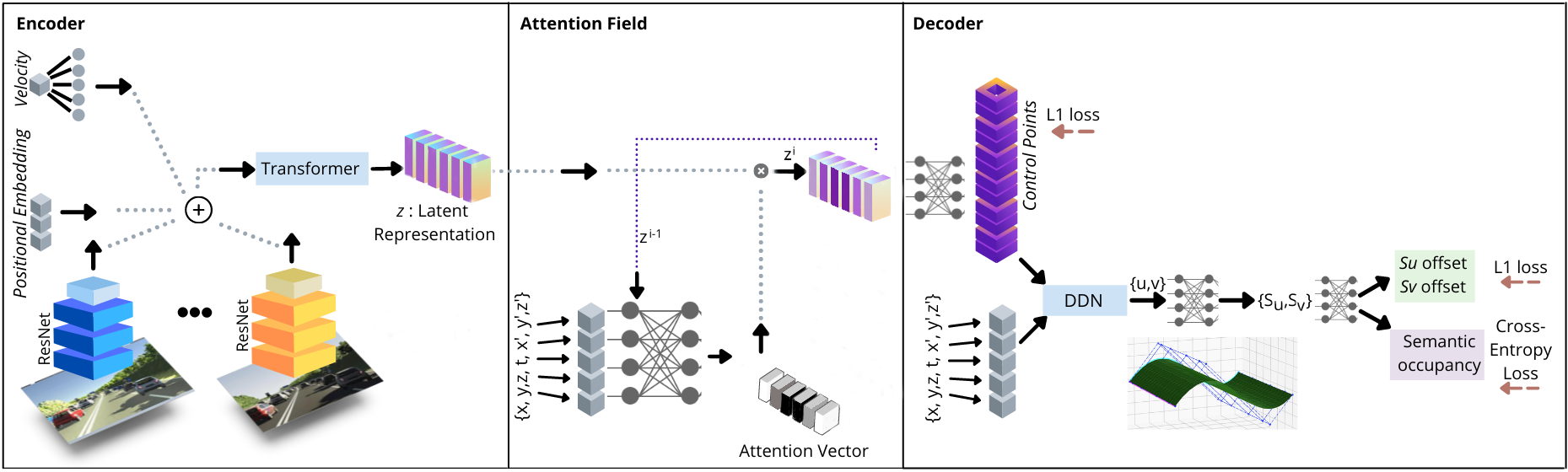}
  \caption{\textbf{Architecture}: 
  The system takes in images from a surround monocular setup, along with the vehicle velocity, and predicts control points, offset and occupancy in the $(s_u,s_v)$ space given a query point $\textbf{q}$ on the manifold. At test time, we sample points from surface to obtain offsets and semantics, which are used to generate drive commands.}
  \label{fig:arch}
\end{figure*}

In Fig. \ref{fig:arch}, motivated by \cite{neat}, we structure our approach to predict waypoints and semantics on the manifold in an end-to-end manner, learning from expert demonstrations. 
% Waypoint history $\textbf{w}_t,~t\in\{1\dots T\}$ form the ego-vehicle trajectory history.
The vehicle coordinate system is defined with position of ego-vehicle being at origin $(0,0,0)$ at current instant. The right handed coordinate system has the front of the vehicle in positive X-axis and Z-axis pointing upwards. The architecture consists of an encoder, attention field and a decoder. 
\begin{figure}[t]
\centering
\includegraphics[width=0.9\linewidth]{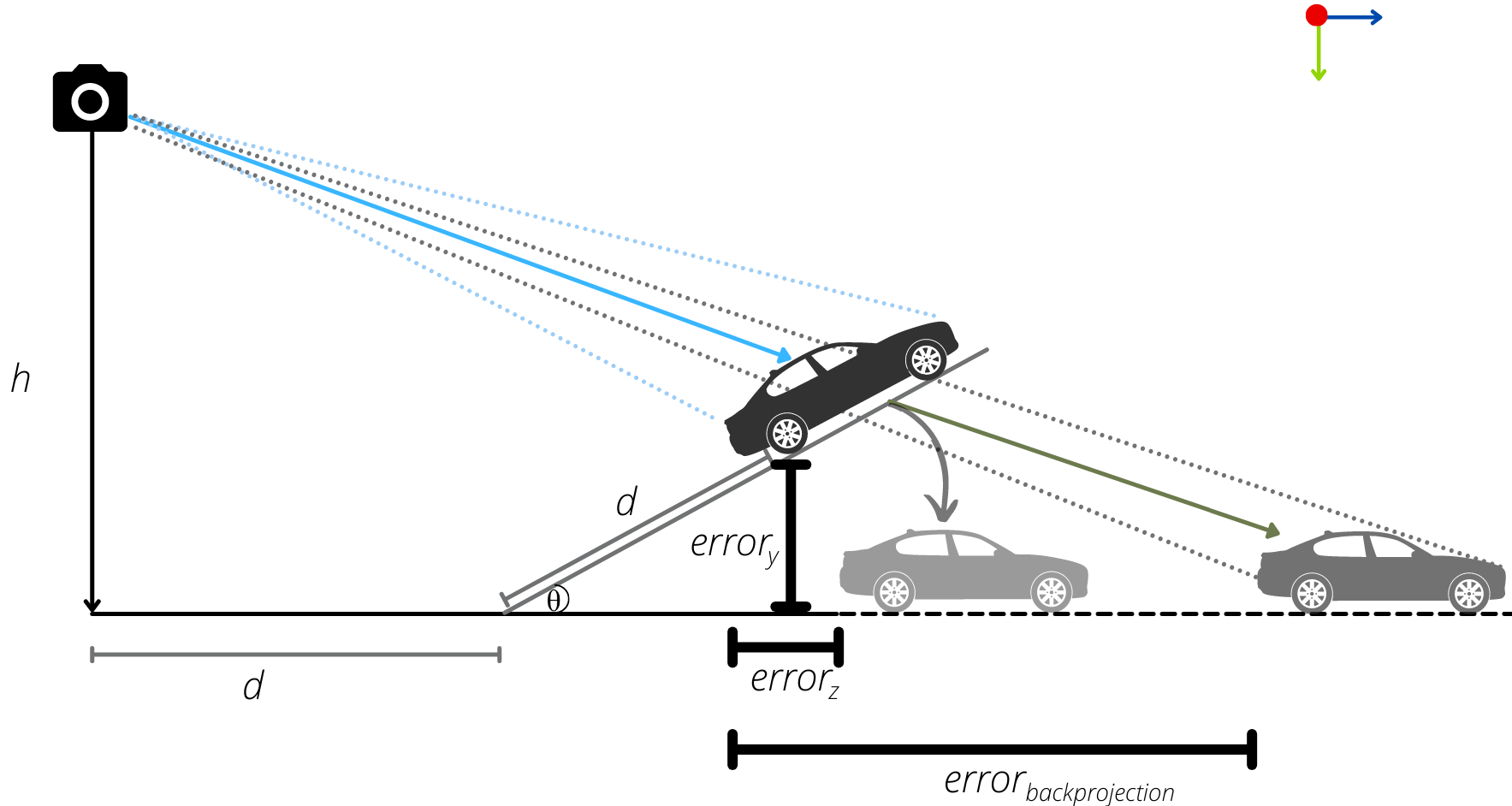}
\caption{Co-planarity assumption when the object is sharing a different planar profile than the ego-vehicle can lead to error in localization of the object, and incorrect inputs to planning.}
\label{fig:whyGroundHelps}
\end{figure}

\textbf{Encoder}: As our agent drives through the scene, we collect sensor inputs $\textbf{X}$ from the surround monocular cameras, over $T$ time steps, where $\textbf{X} = 
\textbf{x}_{s,t}$, $s=1:S$, $t=1:T$, where $S$ is the number of sensors. Each RGB image, $\textbf{x}_{s,t} \in \mathbb{R}^{W\times H\times 3}$ is passed through a Res-Net \cite{resnet} to obtain a feature representation of the image. This, along with the vehicle speed $v \in \mathbb{R}$ and a learned position embedding, is summed and passed through a transformer. The transformer integrates the features globally, adding contextual cues to each patch with its self-attention mechanism. This enables interactions over a large spatial regions and across the different sensor outputs. The output of the transformer is a latent encoding, represented by $\textbf{Z}$ $\in \mathbb{R}^{(S \ast T \ast P) \times C}$  . The encoder operation is depicted in Eq.\ref{eq:encoder} below:
\begin{equation}
\text{Encoder} (\mathcal{E}) : \mathbb{R}^{S \times T \times W \times H \times 3} \times \mathbb{R} \longmapsto \mathbb{R}^{\left(S \ast T \ast P\right) \times C}
\label{eq:encoder}
\end{equation}
where $P=$number of spatial features and $C=$ feature dimensionality.

\textbf{Attention Field}: We define a query point on the manifold $\textbf{q}\in  \mathbb{R}^7 = (x,y,z,t,x',y',z')$, where $t$ is time, $\textbf{q}_{P}:(x,y,z) \in \mathbb{R}^3$ is the query location and $\textbf{q}_{P'}:(x',y',z') \in \mathbb{R}^3$ is the target location. To attend to patch features for a particular query point $\textbf{q}$, we adopt the iterative attention mechanism of \cite{neat}. Specifically, at each iteration $\textit{i}$, the output of the attention field $A_i$ is used to relatively weigh each of the features $\textbf{Z}_i$, based on their specific relevance to a query point $\textbf{q}$. This is used as input of the attention along with  $\textbf{q}$ at the next iteration. For the first iteration, each of the $A_i$'s are initialized with a uniform scalar - signifying an uniform attention to start with. The weights of the attention network are shared across all $\textit{N}$ iterations. The attention mechanism is denoted in Eq.\ref{eq:att} below:
\begin{equation}
\text{Attention} (A) : \mathbb{R}^{7} \times \mathbb{R}^{C} \longmapsto \mathbb{R}^{(S \ast T \ast P)}
\label{eq:att}
\end{equation}
To capture the correlation of the presence of traffic participants and the road profile on which they are present (e.g. a car is on the surface ) a common attention field is proposed. 
\begin{figure}[!ht]
    \begin{subfigure}{0.5\linewidth}
        \centering
        \includegraphics[scale=0.23]{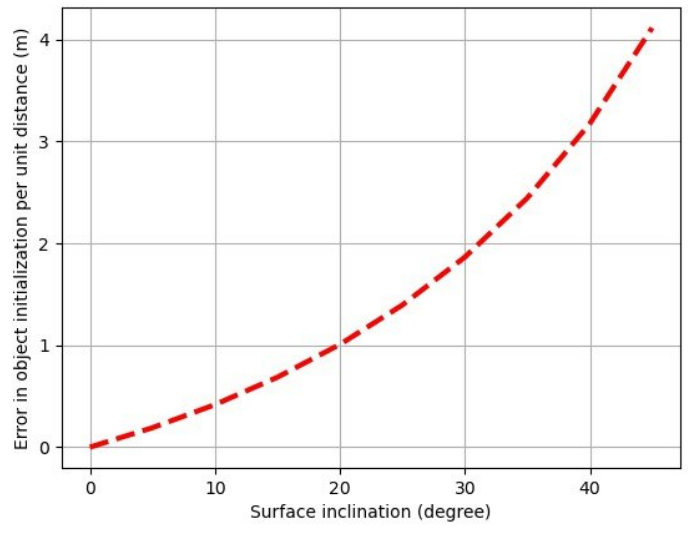}
        \caption{}
        \label{fig:loss_a}
    \end{subfigure}%
    \begin{subfigure}{0.5\linewidth}
        \centering
        \includegraphics[scale=0.23]{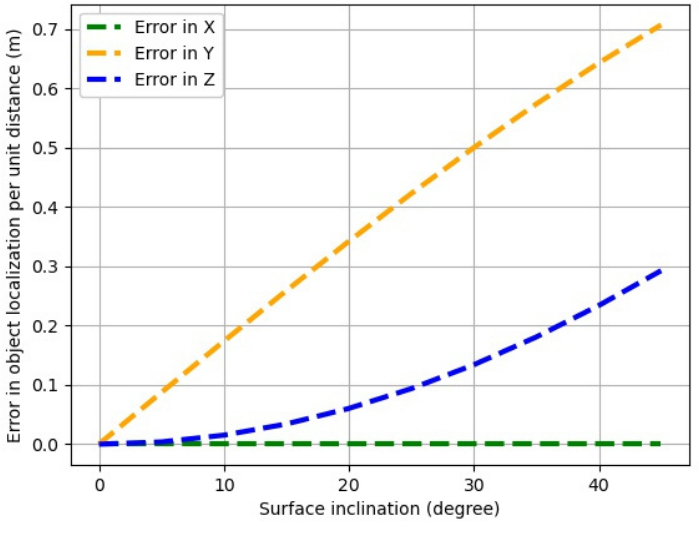}
        \caption{}
        \label{fig:loss_b}
    \end{subfigure}
    \caption{(a) Error in localization ($\text{error}_\text{backprojection}$) of object if co-planarity assumption is enforced on non-planar profiles  \cite{mobileye,chandraker2015},  (b) Coordinate wise error ($\text{error}_\text{X,Y,Z}$) in estimates of position in ego reference frame  when the inclination is neglected. Unit dimensions ($h=1,d=1$) are assumed.}
\end{figure}
\textbf{Decoder}: Given features $\{\textbf{Z}_i$\}, a grid of $E\ast F$ control points $ \{\textbf{P}_c^{ef}\}_{e \in [1, E], f \in [1 \dots F]}$ that govern the manifold representation of the scene are extracted from a Multi Layer Perceptron(MLP). Each control point $\textbf{P}_c^{ef}\in \mathbb{R}^3$, which makes the output of MLP $\{\textbf{P}_c^{ef}\}\in \mathbb{R}^{(E \ast F) \times 3}$. Given $\textbf{q}_P:(x,y,z) \in \mathbb{R}^3$ and the the control points $\{\textbf{P}_c^{ef}\} \in \mathbb{R}^{(E\ast F) \times 3}$, we use a Deep Declarative Network(DDN)  \cite{ddn} layer to obtain the reverse mapping $\textbf{q}_P\in\mathbb{R}^3 \longmapsto (u,v)\in\mathbb{R}^2$ by minimizing the Eq.\ref{eqn:minbspline} below, obtained by rearranging Eq.\ref{eqn:bspline}:
\begin{equation}
\label{eqn:minbspline}
% \text{arg}\,\text{min} \limits_{0\leq u,v\leq 1}\
\arg\min_{u,v}
\left|\textbf{q}_P- \sum_{e=1}^{E} \sum_{f=1}^{F} \textbf{P}_{c}^{ef} \cdot B_{e}^E(u)\cdot B_{f}^F(v)\right|þ, \text{s.t. }~0\leq u,v \leq
\end{equation}
Subsequently, we remap the $(u,v) \in \mathbb{R}^2$ space to isometric arch length space $(s_u,s_v) \in \mathbb{R}^2$ through an MLP. Next, the decoder predicts the semantic class $s_i \in \mathbb{R}^M$ (where M is the number of classes) and waypoint offset $o_i \in \mathbb{R}^2$ at each of the N attention iterations. 
% The architecture is shown in Fig. \ref{fig:arch}
\section{Training and Inference}
At train time, the outputs from each of the iterations are supervised to assist with convergence of loss. The predicted control points $\{\textbf{P}_c^{ef}\} \in \mathbb{R}^{(E\ast F) \times 3}$ are supervised via $\mathcal{L}_1$ loss function. For the semantic and offset prediction in the $(s_u,s_v)$ space, Cross-Entropy(CE) loss and $\mathcal{L}_1$ losses are used respectively. The final loss function takes the form as shown in Eq.\ref{eqn:loss_total} below:
\begin{equation}
\centering
\begin{aligned}
 \mathcal{L}_{total} = \eta_{PC}\mathcal{L}_{PC} + \eta_{Off}\mathcal{L}_{Off} + \eta_{CE}\mathcal{L}_{CE}
\end{aligned}
\label{eqn:loss_total}
\end{equation}
where, $\eta_{PC}$, $\eta_{Off}$, $\eta_{CE}$ are weighing factors that control the relative importance of each of the loss terms. We predict both observed ${t=1...T}$ and the future ${t=T+1,...T+H_z}$ semantic information to have a more holistic understanding of the scene. For each of the $M$ semantic classes, $K$ points are sampled in an edge aware manner to accurately capture the semantic boundaries. A coverage loss is also imposed to sampling to ensure the SOG space is uniformly covered as much as possible.At test time, the surround monocular images and a query point $\textbf{q}$ is the input. The semantic occupancy and offsets are obtained at the end of $N^{th}$ attention iteration. 

\section{Experiments}
In Fig.\ref{fig:teaser} we show visualization of a scene from Carla \cite{carla} and how ignoring the road gradients while formulation the motion planning in BEV can lead to incorrect inference. In Fig. \ref{fig:whyGroundHelps} we show how ignoring of ground plane topology leads to incorrect object localization. In  Fig. \ref{fig:loss_a} and Fig \ref{fig:loss_b}, we show the error in localization and inputs to planning as the plane inclination changes. We plan to carry out experiments in Carla \cite{carla} and SYNTHIA-SF \cite{synthia-sf} which contain scenarios with different road-profiles and gradients. The experiments aim to predict accurate SOG and GOF. The proposed approach would be compared with \cite{codevilla2019exploring,lbc,prakash2021multi,chen2021learning,toromanoff2020end} while evaluating on Carla and  \cite{ansari2018earth} while evaluation on SYNTHIA-SF \cite{synthia-sf}. It has to be noted that for \cite{ansari2018earth} only deals with the task of object localization in 3D, hence semantic occupancies would need to be lifted to 3D bounding boxes in order to have a fair comparison. For experiments, $P=64, C=512, H_z=4, E \ast F = 35, K = 128, M = 5, T = 1, N =2$. The semantic classes that we consider are \textit{\{none, road, obstacle, red-light, green-light\}}.

\section{Conclusion}
We present NMR, an approach that captures the accurate road surface for the task of end-to-end autonomous driving. We lift the raw sensor input data and vehicle state to a high dimensional latent embedding. Attention fields are used to extract control points that govern the surface geometry, and semantic occupancy and offset information given any query point in $\mathbb{R}^3$ on the manifold. We present edge-aware sampling methods to accurately capture the occupancy information in the scene. We propose to test our approach on challenging road topologies in Carla\cite{carla} and SYNTHIA-SF\cite{synthia-sf}.

\bibliographystyle{IEEEtran}  
\bibliography{references.bib}
\end{document}